\documentclass[a4paper, 10pt, conference]{ieeeconf}      

\IEEEoverridecommandlockouts             

\usepackage{graphicx}
\graphicspath{{../figures}} 
\usepackage{hyperref}
\hypersetup{
    colorlinks=false,
    pdfborder={0 0 0}
}

\title{\LARGE \bf
Towards a Simple Framework of Skill Transfer Learning for \\ Robotic Ultrasound-guidance Procedures 
}

\author{Tsz Yan Leung$^{1}$ and Miguel Xochicale$^{2}$
\thanks{$^{1}$
	King's College London, UK
       {\tt\small tsz\_yan.leung@kcl.ac.uk}}%
\thanks{$^{2}$
	Currently University College London, UK. 
        Previously King's College London, UK.
        {\tt\small m.xochicale@ucl.ac.uk}}%
}

\begin{document}
\maketitle
\thispagestyle{empty}
\pagestyle{empty}

\begin{abstract}
In this paper, we present a simple framework of skill transfer learning for robotic ultrasound-guidance procedures.
We briefly review challenges in skill transfer learning for robotic ultrasound-guidance procedures.
We then identify the need of appropriate sampling techniques, computationally efficient neural networks models that lead to the proposal of a simple framework of skill transfer learning for real-time applications in robotic ultrasound-guidance procedures.
We present pilot experiments from two participants (one experienced clinician and one non-clinician) looking for an optimal scanning plane of the four-chamber cardiac view from a fetal phantom.
We analysed ultrasound image frames, time series of texture image features and quaternions and found that the experienced clinician performed the procedure in a quicker and smoother way compared to lengthy and non-constant movements from non-clinicians.
For future work, we pointed out
the need of pruned and quantised neural network models
for real-time applications in robotic ultrasound-guidance
procedure.
The resources to reproduce this work are available at \url{https://github.com/mxochicale/rami-icra2023}.
\end{abstract}


\section{INTRODUCTION}
Ultrasound (US) imaging is a popular imaging modality because of its affordability, 
non-ionising imaging, and real-time capabilities.
Recently, the field of ultrasound-guidance procedures has been advanced with the development of robotic ultrasound systems that range from tele-operated, semi-autonomous and fully autonomous~\cite{deng2021, vonHaxthausen2021, Gerlach2022}. 
However, there are still scientific and technical challenges in robotic ultrasound-guidance procedures: 
(a) traditional imaging is user-dependent, skill-dependant and device-dependent \cite{chen1997},
(b) traditional hardware for human motion tracking is usually expensive or cumbersome~\cite{Dressler2021}, and 
(c) frameworks are designed for specific types of sensors, clinical US devices, robots and operating systems~\cite{niu2022}.
Learning ultrasound skills from sonographers that look for the optimal scanning plane (OSP) is still an open challenge in robotic ultrasound-guidance procedures~\cite{deng2021}.
It is hypothesised that robotic ultrasound-guided procedures would require to be simple, less expensive and less cumbersome for skill transfer learning.
Hence, in this paper, we are proposing a simple framework of skill transfer learning for robotic ultrasound-guidance procedures.
This paper is divided into robotic ultrasound-guidance procedures, framework for sonographer skill transfer learning, results and conclusions with future work.

\section{ROBOTIC ULTRASOUND-GUIDANCE PROCEDURES}
Robotic ultrasound systems are actively investigated for teleoperatation, semi-autonomous and fully autonomous modalities~\cite{vonHaxthausen2021}.
For instance, Deng et al. proposed a multi-modal task learning architecture for ultrasound scanning skills with input data from ultrasound images, force and probe pose, stating the challenge of real-time guidance due to computational signal processing~\cite{deng2021}.
Robotic US-guidance radiation therapy for lesion in abdomen has been successful using CNN-based search.
However, treatment plans without US-robot guidance are still superior to robotic US-guidance because of the quality of acquired ultrasound images~\cite{Gerlach2022}.

\begin{figure}[t]
\centering
\includegraphics[width=0.47\textwidth]{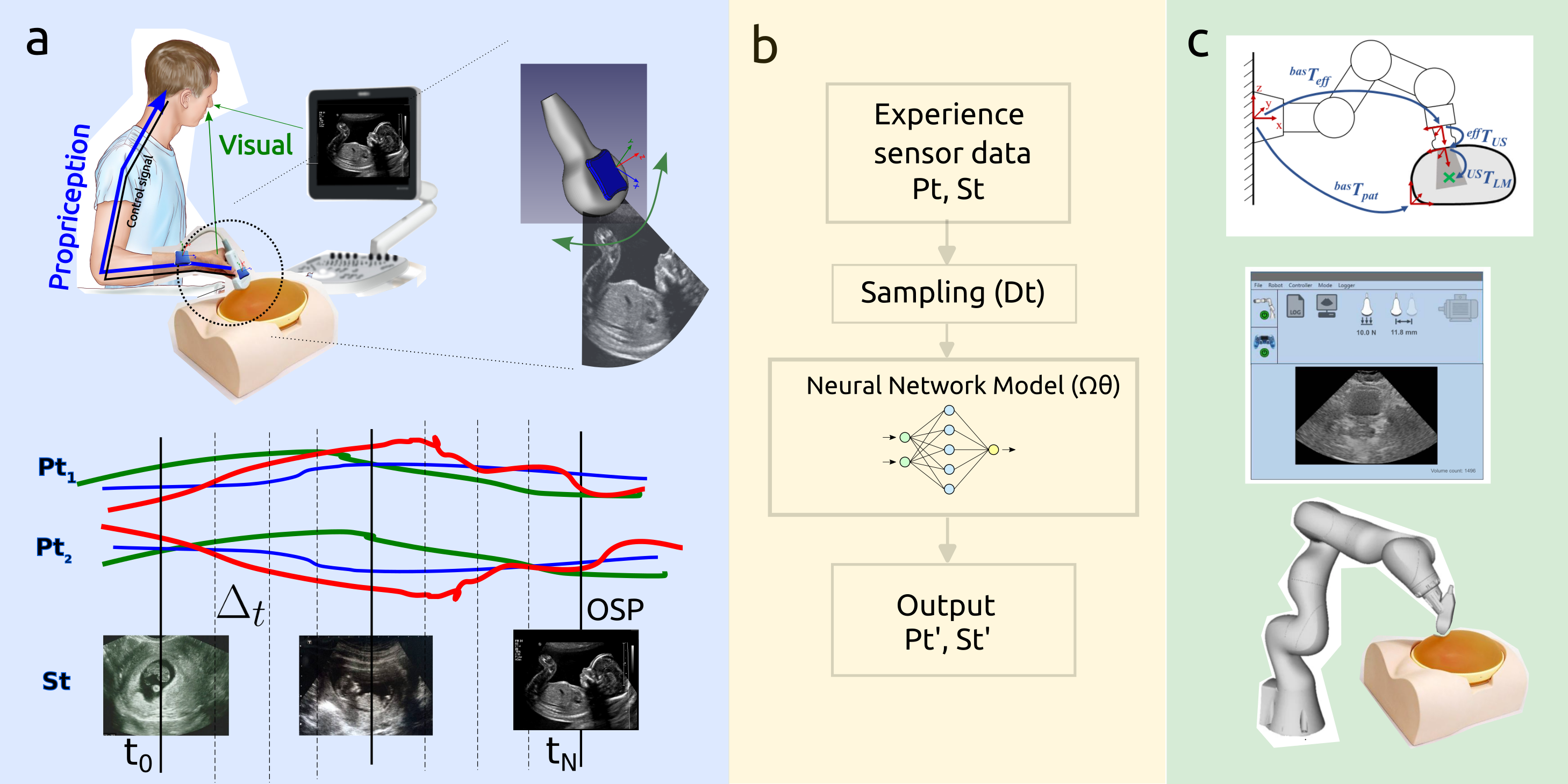} 
    \caption{
            \textbf{(a)} Ultrasound-guidance procedures:
		sonographer operating an ultrasound machine with fetal phantom and sensor fusion signals from inertial sensors and ultrasound imaging;
            \textbf{(b)} Simple framework for skill transfer learning: 
		collecting experience with sensors ($Pt_n$ pose and $St$ Signal), sampling method for fusion sensor ($\Delta_t$), and identified the need of computational efficient neural network model ($\Omega_\theta$), and output for high-dimensional model ~\cite{deng2021}, and 
            \textbf{(c)} Robotic ultrasound-guidance procedures:
		transformations, graphical user interface and simulation using robotic US-guidance light-weight 7 degrees-of-freedom robot (KUKA LBR Med 7)~\cite{Gerlach2022, Ipsen2021}.
       }
\label{fig:main}
\end{figure}


\section{FRAMEWORK FOR SONOGRAPHER SKILL TRANSFER LEARNING}
Modelling the optimal scanning plane (OSP) requires ultrasound images, spacial locations and anatomical understanding from experienced sonographers (Fig~\ref{fig:main}a)~\cite{deng2021,vonHaxthausen2021}.
Current systems to model OPS are expensive and cumbersome leading to the need of smaller and low-cost systems~\cite{Dressler2021}. 
One potential avenue to reduce cost and ergonomic form factor is (a) the use of Inertial Measurement Units (IMU) to track probe position during US scanning procedures~\cite{PREVOST2018187}
and (b) the application of the appropriate fusion techniques of ultrasound video signal and motion signal from IMU for probe guidance~\cite{droste2020}, and (c) computationally efficient neural networks models~\cite{deng2021}.
Hence, we introduce a framework for sonographer skills transfer learning, considering: sampling technique and fusion sensor techniques (IMU and US) and discuss the need of real-time guidance with  pruned and quantised neural network models, feature extraction and transfer learning for robotic-ultrasound-guidance procedures (Fig~\ref{fig:main}b, c).





\section{EXPERIMENTS: DESIGN AND RESULTS}
\subsection{Experiment setup}
Considering fetal biometry parameters from the NHS 20-week screening scan protocol and National Health Service (NHS) foetal anomaly screening programme (FASP)~\cite{NHS_england2022},
we conducted a pilot experiment with two participants (one non-clinical and one clinical with 10 years of experience in echocardiography).
During the experiment, we asked participants to find the optimal scanning plane for the four chamber view of foetal ultrasound examination phantom ("SPACE FAN-ST", Kyoto Kagaku Co., Ltd, Kyoto, Japan).
We attached inertial sensors (LPMS-B2, LP-Research Inc., Tokyo, Japan) to a convex US probe to track the pose of the ultrasound image from a clinical ultrasound device (EPIQ 7G, Koninklijke Philips N.V., Amsterdam, Netherlands) with the use of a CPU laptop computer that streamed images via a USB framegrabber (Mirabox).

\subsection{Experimental results}
\begin{figure}[t]
\centering
\includegraphics[width=0.44\textwidth]{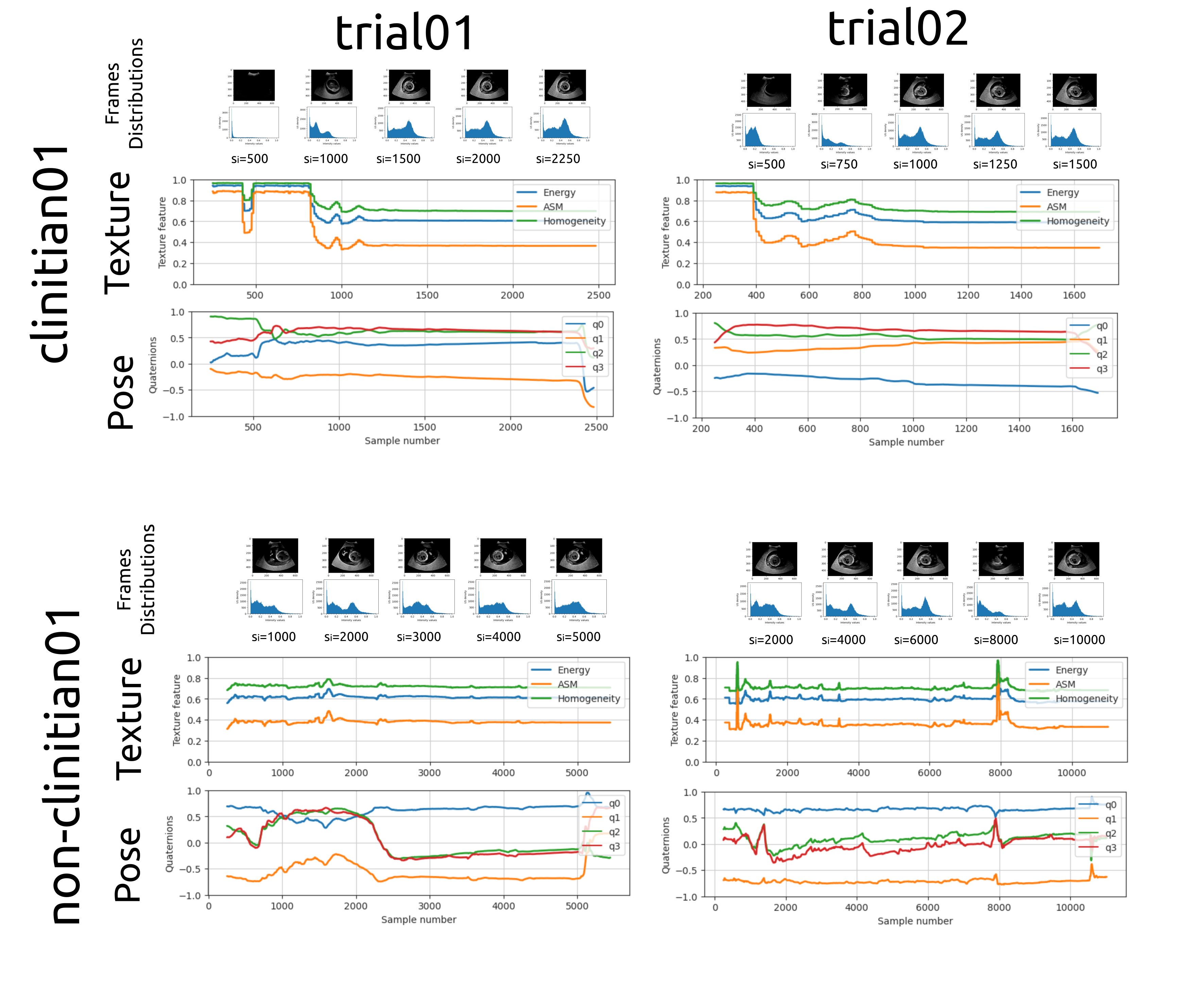} 
\caption{
Ultrasound image frames, image distributions, image texture features (energy, Angular Second Momentum ASM, and homogeneity) and pose time-series (quaternions) from one experienced clinician (clinician01) and one non-clinician (non-clinician01) in two trials (trial01 and trial02), looking for four-chamber optimal scanning plane. 
       } 
\label{fig:results}
\end{figure}

Fig~\ref{fig:results} shows the time series from one experienced clinician and one non-clinician while looking for an optimal scanning plane of the four-chamber view.
It can be noted that the expert took less time (1600 and 2500 samples) to find an optimal scanning plane whereas non-clinician took a greater time (5000 and 10000 samples).
Similarly, the time-series for the experienced clinician appear to be more smooth and consistent compared to the jerkiness of non-clinical participant. 

 
\section{CONCLUSIONS AND FUTURE WORK}
We have presented a simple framework of skill transfer learning for robotic ultrasound-guidance procedures.
We presented sensor fusion methods and sampling rate techniques for optimal scanning plane of the four-chamber view from two participants (one experienced clinician and one non-clinicians), where experienced clinician showed an smother and quicker procedure compare to a lengthy and non-constant movement of non-clinician.
For future work, we pointed out the need of pruned and quantised neural network models for real-time applications in robotic ultrasound-guidance procedure. 


\addtolength{\textheight}{-12cm}   


\section*{ACKNOWLEDGMENT}
Thanks to Tsz Yan Leung for her excellent research work  during her M.Sc. in Medical Physics and Engineering in 2022 at King's College London.
Thanks to Nhat Phung for volunteering as experienced sonographer in the experiments to identify optimal scanning plane of fetal four-chamber views at St Thomas' Hospital. 

\bibliographystyle{IEEEtran}


\end{document}